\documentclass{article} 
\usepackage{iclr2023_conference_tinypaper,times}

\usepackage{amsmath,amsfonts,bm}

\def\eqref#1{equation~\ref{#1}}

\def\1{\bm{1}}

\DeclareMathAlphabet{\mathsfit}{\encodingdefault}{\sfdefault}{m}{sl}
\SetMathAlphabet{\mathsfit}{bold}{\encodingdefault}{\sfdefault}{bx}{n}

\usepackage{url}

\usepackage{graphicx}
\usepackage{subfigure}
\usepackage{booktabs} 
\usepackage{siunitx}

\usepackage{booktabs}
\usepackage{caption}
\usepackage{colortbl}
\usepackage{multirow}
\usepackage{amsmath}

\usepackage{rotating}
\usepackage{wrapfig}

\definecolor{darkblue}{RGB}{25, 50, 112}
\usepackage[colorlinks=true, linkcolor=black, citecolor=darkblue]{hyperref}
\definecolor{ROW_COLOR}{HTML}{C9F7F4}
\definecolor{COLOR_MEAN}{HTML}{f0f0f0}

\title{
Empirical Study on Updating Key-Value Memories in Transformer Feed-forward Layers
}

\author{$^{1}$$^{2}$Zihan Qiu\thanks{Equal contribution} $^{ }$ \thanks{Work done while interning at HKUST.} $^{ }$,  $^{3}$Zeyu Huang$^*$,  $^{4}$Youcheng Huang$^*$,  $^{1}$Jie Fu\thanks{Corresponding author}  \\
$^{1}$CSE, HKUST $\,\,\,$ $^{2}$IIIS, Tsinghua University, \\
$^{3}$ILCC, University of Edinburgh  $\,\,\,$  $^{4}$College of Computer Science, Sichuan University   \\
\texttt{qzh11628@gmail.com, zeyu.huang@ed.ac.uk} \\
\texttt{youchenghuang@stu.scu.edu.cn, jiefu@ust.hk} \\
}

\iclrfinalcopy 
\begin{document}

\maketitle

\begin{abstract}
The feed-forward networks (FFNs) in transformers are recognized as a group of key-value neural memories to restore abstract high-level knowledge. 
In this work, we conduct an empirical ablation study on updating keys (the 1st layer in the FFNs layer) or values (the 2nd layer in the FFNs layer). 
We compare those two methods in various knowledge editing and fine-tuning tasks of large language models to draw insights to understand FFNs further. Code is available at \href{https://github.com/qiuzh20/Tuning-keys-v.s.-values}{this repo}.

\end{abstract}

\section{Introduction}




How do pre-trained Transformer models process and store information? 
~\citet{DBLP:conf/emnlp/GevaSBL21,DBLP:conf/emnlp/GevaCWG22} suggest feed-forward networks (FFNs) operate as key-value neural memories~\citep{DBLP:conf/nips/SukhbaatarSWF15}. 
Specifically, given hidden states $h\in \mathbb{R}^{d_e}$, where $d_e$ is embedding size, $\text{FFNs}(h)=f(h\cdot K^T)\cdot V$, where $K, V \in \mathbb{R}^{d_m \times d_e}$, $d_m$ is the width of FFNs and $f$ is the non-linear activation. 
Each row of $K$ can be viewed as a key correlated with input textual patterns, and each row of $V$ can be viewed as a value that induces a distribution shift over the residual stream.
For example, when a model predicts the next token based on the prefix `Eiffel Tower is located in', $k_2$ is activated so that $v_2$ can promote the probability of `Paris' in the output. On the contrary, values for irrelevant concepts (e.g., $v_1$ for `Cat', $v_d$ for Soccer) are deactivated and can not dominate the output distribution.

\begin{figure}[h]
    \vspace{-1.05em}
    \centering
    \includegraphics[width=0.68\columnwidth]{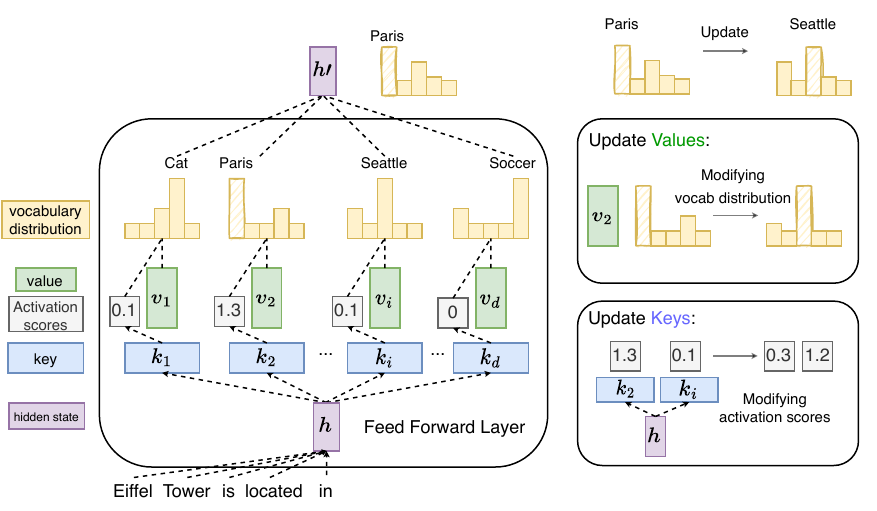}
    \vspace{-1em}
    \caption{\footnotesize 
\textbf{Left}: FFNs operation is conceptualized as a key-value memory. The input hidden states interact with \textit{keys} (rows of \( K \)) through an inner product to obtain activation values. These activations then serve as weights for summing \textit{values} (rows of \( V \)). 
\textbf{Right}: To update these key-value memories, one can either directly modify the values that store relevant information or alter the keys to adjust the weights of existing values.}
    \label{fig:key_tuning}
    \vspace{-0.8em}
\end{figure}

Based on this point of view, how can we update the information processing and storing? We employ Knowledge Editing (KE)~\citep{wang2023knowledge} as an illustrative example: altering `Paris' to `Seattle' in response to `Eiffel Tower is located in' by modifying the model weights.
In this work, we compare two updating choices as shown in Figure~\ref{fig:key_tuning}:
1. \textbf{Values}: 
tuning $v_2$ to shift its concept from `Paris' to `Seattle'.
2. \textbf{Keys}: 
tuning $k_2$ to boost the activation of $v_i$ for the target concept `Seattle'.

We test the two methods in various scenarios for different pre-trained transformers, including knowledge editing~\citep{DBLP:conf/emnlp/CaoAT21, DBLP:conf/iclr/HuangSZZR023}, 
multi-task tuning~\citep{DBLP:conf/iclr/AribandiTSRZMZ022}, and instruction-tuning~\citep{DBLP:conf/iclr/WeiBZGYLDDL22}. 
We generally recognize the superiority of updating keys over updating values.
We contend that \textit{compared to directly modifying the model's knowledge (values), altering the mechanism of controlling this knowledge (keys) can be more effective.}

\begin{table}[t!]
\small
    \vspace{-4em}
    \setlength{\tabcolsep}{4pt}
    \renewcommand{\arraystretch}{1.0}
    \centering
    \caption{\footnotesize Knowledge Editing on GPT-J (6B)~\citet{mesh-transformer-jax}}
    \vskip -0.1in
    \label{tab:editing_target_efficacy}
    \resizebox{0.7\textwidth}{!}{
    \begin{tabular}{l|ccccc}
    \toprule
    \rowcolor{COLOR_MEAN}
    \textbf{Editing Target} & \textbf{Efficacy}\,$\uparrow$ & \textbf{Paraphrase}\,$\uparrow$ & \textbf{Specificity}\,$\uparrow$ & \textbf{Score}\,$\uparrow$ & \textbf{Time\,(s)}\,$\downarrow$ \\
    \midrule
    \multicolumn{6}{c}{1 Counterfact Editing on GPT-J (6B)} \\
    \midrule
    On Value & 100.00 & 98.18 & 6.04 & 16.15 & 0.79 \\
    \rowcolor{ROW_COLOR}
    On Key & 100.00 & \textbf{98.44} & \textbf{28.89} & \textbf{54.77} & 0.83 \\
    \midrule
    \multicolumn{6}{c}{1 zsRE Editing on GPT-J (6B)} \\
    \midrule
    On Value & 99.11 & 56.32 & 21.81 & 40.71 & 83.12 \\
    \rowcolor{ROW_COLOR}
    On Key & 98.57 & \textbf{69.19} & \textbf{24.64} & \textbf{46.02} & \textbf{12.63} \\
    \bottomrule
    \end{tabular}
    }
    \vskip -0.2in
\end{table}

\section{Experiments}

\textbf{Knowledge Editing} is to edit specific information in the model while leaving irrelevant ones uninfluenced. In KE, knowledge refers to the triplets of (subject, relationship, object).
The editing is done by maximizing the probabilities of the object tokens given a prompt containing (subject, relationship).
We follow the same experiment settings as~\citet{DBLP:conf/iclr/MengSABB23},
more details can be found in Appendix~\ref{sec:rep_edit}.
We simply introduce the evaluation metrics:
\textbf{Efficacy} measures the editing success,
\textbf{Paraphrase} measures the editing generalization in different but related contexts,
\textbf{Specificity} measures the editing locality in unrelated contexts, and \textbf{Score} aggregates the three metrics by taking the harmonic mean.
\citet{DBLP:conf/iclr/MengSABB23} solves a constrained linear problem for model updating.
But the same method cannot update $K$ because of the non-linear $f(\cdot)$.
We thus update $K$ and $V$ through back-propagation to examine their performances fairly.
Table~\ref{tab:editing_target_efficacy} reports the results of updating $K$ or $V$.
While updating each component gives $100\%$ efficacy, updating $K$ shows more generalization and locality, achieving a large performance gain on \textbf{Score}.
Notably, updating $V$ suffers from much higher time-cost, indicating the hardness of shifting models `concept' compared with updating the `concept' usage.
Reproducibility details and more results are in Appendix~\ref{sec:rep_edit} and~\ref{sec:more_edit}.

\begin{table}[htbp!]
\small
    \vspace{-1.1em}
    \setlength{\tabcolsep}{4pt}
    \renewcommand{\arraystretch}{1.0}
    \centering
    \caption{\footnotesize 
    LoRA instruction and multi-task tuning results.
    }
    \vskip -0.1in
    \label{tab:lora_performance}
    \resizebox{0.85\textwidth}{!}{
    \begin{tabular}{l|c|c|ccccccc}
    \toprule
    \rowcolor{COLOR_MEAN}
    \textbf{Lora Target} & \textbf{Trainable} & \textbf{MMLU} & \textbf{Bool-q} & \textbf{CB} & \textbf{20-news} & \textbf{Race} & \textbf{COPA} & \textbf{QNLI} & \textbf{6-Avg} \\
    \midrule
    \multicolumn{10}{c}{LoRA rank=8 on Llama2-7B} \\
    \midrule
    q v & 0.062\% & 46.77 & 58.20 & 46.43 & 61.50 & 60.50 & 73.00 & 57.90 & 59.59 \\
    $\text{Value}_{\text{down}}$ & 0.057\% & 46.28 & 60.70 & 53.57 & 67.40 & 60.90 & 82.00 & 53.50 & 63.01 \\
    $\text{Key}_{\text{gate}}$ & 0.057\% & 46.79 & 69.00 & 64.29 & 55.30 & 63.40 & 79.00 & 77.80 & 68.13 \\
    \rowcolor{ROW_COLOR}
    $\text{Key}_{\text{up}}$  & 0.057\% & \textbf{46.99} & 74.10 & 58.93 & 56.80 & 62.50 & 79.00 & 78.40 & \textbf{68.29} \\
    \midrule
    q v $\text{Value}_{\text{down}}$ & 0.119\% & 46.92 & 66.30 & 48.21 & 59.10 & 64.80 & 68.00 & 62.30 & 61.45 \\
    q v $\text{Key}_{\text{gate}}$ & 0.119\% & 46.99 & 59.20 & 62.50 & 61.20 & 66.10 & 69.00 & 63.10 & 63.52 \\
    \rowcolor{ROW_COLOR}
    q v $\text{Key}_{\text{up}}$  & 0.119\% & \textbf{47.13} & 64.20 & 51.79 & 66.00 & 65.00 & 68.00 & 68.80 & \textbf{63.96} \\
    \midrule
    \multicolumn{10}{c}{LoRA rank=16 on Llama2-7B} \\
    \midrule
    q v & 0.124\% & 46.9 & 68.30 & 51.79 & 65.00 & 63.60 & 76.00 & 52.70 & 62.90 \\
    $\text{Value}_{\text{down}}$ & 0.115\% & 46.75 & 67.80 & 41.07 & 54.40 & 62.70 & 71.00 & 61.90 & 59.81 \\
    $\text{Key}_{\text{gate}}$ & 0.115\% & 46.91 & 71.40 & 55.36 & 57.00 & 62.90 & 79.00 & 76.60 & 67.04 \\
    \rowcolor{ROW_COLOR}
    $\text{Key}_{\text{up}}$ & 0.115\% & \textbf{47.02} & 72.60 & 62.50 & 66.70 & 61.60 & 81.00 & 64.70 & \textbf{68.18} \\
    \bottomrule
    \end{tabular}}
    \vskip -0.07in
\end{table}

\textbf{Instruction and Multi-task Tuning.}
We extend experiments to a more general LoRA-tuning of the Llama2-7B~\citep{touvron2023llama} model. 
This involves separately: 1. tuning the model on the Alpaca dataset~\citep{alpaca} and then testing on MMLU~\citep{hendrycks2020measuring}. 
2. multi-task tuning on six diverse tasks and testing on their respective test sets.
Given that Llama2's FFNs employ SwiGLU~\citep{DBLP:journals/corr/abs-2002-05202}, formulated as \(\text{FFN}_{\text{SwiGLU}} = (\text{Swish}(x \cdot K_{\text{gate}}^T) \otimes (x \cdot K_{\text{up}}^T)) \cdot V_{\text{down}}\), updating to the \(K\) correspond to changes in the $K_{\text{gate}}$ and $K_{\text{up}}$, while changes to the $V$ pertain to $V_{\text{down}}$. 
We explored combinations of LoRA weight additions in the standard attention's q and v, and in the three weighs of FFNs.
More settings and reproducibility can be found in the Appendix~\ref{sec:rep_tuning}.
According to Table~\ref{tab:lora_performance}, across various settings (LoRA rank=8,16; with and without tuning q and v), the performance of LoRA on $\text{Key}_{\text{gate}}$ and $\text{Key}_{\text{up}}$ is significantly better than on $\text{Value}_{\text{down}}$.
More consistent experiment results with 4-bit quantization can be found in Appendix Table~\ref{tab:lora_performance_4bit}.

\section{Conclusion}

We empirically find that updating keys within FFNs yields better performance than updating values when tuning LLMs.
One possible reason could be updating keys solely involves changing the inner product between the keys and given hidden states. 
In contrast, updating values requires accurate optimization corresponding to the intended update.
These characteristics also translate into better results when LoRA-tuning keys instead of values.
It's crucial to note that our objective isn't to propose a better tuning method, so some results may not differ significantly. Instead, we hope
the experiment provides insight for updating pre-trained LLMs: updating the mechanism of how the model controls the knowledge may be more effective than directly modifying the knowledge itself.

\section{URM statement}
The authors acknowledge that at least one key author of this work meets the URM criteria of ICLR 2024 Tiny Papers Track.

\bibliography{iclr2023_conference_tinypaper}
\bibliographystyle{iclr2023_conference_tinypaper}

\appendix
\section{Appendix}

\subsection{More Details for Knowledge Editing and LoRA Tuning}

\textbf{Knowledge Editing}
\label{sec:rep_edit}
Task is to edit specific information in the model while leaving irrelevant ones uninfluenced. In knowledge-editing, knowledge refers to the triplets of (subject, relationship, object). For example, as shown in Fig~\ref{fig:key_tuning}, the knowledge is (`Eiffel Tower', `locates in', `Paris') and the editing is to change the object from the existing one to a new one, i.e., from `Paris' to `Seattle'. 
The editing is done by maximizing the probabilities of the object tokens given a language prompt containing the information of (subject, relationship), e.g., `Eiffel Tower is located in'.

The most difficult part is the data preprocessing and the evaluation.
For these experimental settings, we follow the same processions and codebase\footnote{\url{https://github.com/kmeng01/memit}} as~\citet{DBLP:conf/iclr/MengSABB23}.
While~\citet{DBLP:conf/iclr/MengSABB23} solves a linear problem to update $V$, we update $K$ and $V$ with back-propagation for a fair comparison.
Back-propagation makes our results simple to implement.
We freeze all other modules and update the $K$ or $V$ in the selected layers to update.
We adopt the Adam optimizer with a learning rate of $5e-4$ and weight decay of $0.5$.

\textbf{LoRA Tuning}
\label{sec:rep_tuning}
Our Instruction tuning code is based on the code base\footnote{\url{https://github.com/georgian-io/LLM-Finetuning-Hub/tree/main/llama2}}. 
We built upon the original code by incorporating multi-task Instruction tuning, details of which can be found in our submitted code. 
All training hyperparameters (including training steps) were directly adopted from the original repository, ensuring consistency across all experiments except for differences in LoRA targets.

We utilized the 20-news group dataset for a 20-class news classification task, provided by the original repository. 
For a broader task diversity, we selected Bool-q, CB, and COPA from the SuperGLUE benchmark and QNLI from the GLUE benchmark, along with the QA task Race. 
It's important to note that the sizes of these chosen datasets vary significantly. 
Therefore, we limit each dataset to a maximum of 5000 samples in constructing the multi-task training dataset. This limitation may result in lower performance on some tasks than standard training approaches.

\subsection{More experimental results about the Knowledge Editing}
\label{sec:more_edit}

We conduct more experiments about knowledge editing, especially in editing multi-counterfacts using GPT2-XL.
The reason for not editing multi-counteracts on GPT-J (6B) is that it runs out of memory.
Again, we gently refer readers to~\citet{DBLP:conf/iclr/MengSABB23} for the detailed experiment settings.

\begin{table}[htbp!]
\small
    \vspace{-1.0em}
    \setlength{\tabcolsep}{4pt}
    \renewcommand{\arraystretch}{1.0}
    \centering
    \caption{\footnotesize Results of the Knowledge Editing}
    \vskip -0.07in
    \label{tab:more_edit}
    \resizebox{0.7\textwidth}{!}{
    \begin{tabular}{l|ccccc}
    \toprule
    \rowcolor{COLOR_MEAN}
    \textbf{Editing Target} & \textbf{Efficacy}\,$\uparrow$ & \textbf{Paraphrase}\,$\uparrow$ & \textbf{Specificity}\,$\uparrow$ & \textbf{Score}\,$\uparrow$ & \textbf{Time\,(s)}\,$\downarrow$ \\
    \midrule
    \multicolumn{6}{c}{1 Counterfact Editing on GPT2-xl} \\
    \midrule
    On Value & 100.00 & 71.02 & 68.33 & 77.49 & 1.21 \\
    \rowcolor{ROW_COLOR}
    On Key & 100.00 & \textbf{83.78} & 66.16 & \textbf{80.97} & \textbf{0.97} \\
    \midrule
    \multicolumn{6}{c}{10 Counterfacts Editing on GPT2-xl} \\
    \midrule
    On Value & 100.00 & 45.50 & 66.74 & 63.88 & 49.92 \\
    \rowcolor{ROW_COLOR}
    On Key & 100.00 & \textbf{57.14} & \textbf{73.48} & \textbf{72.97} & \textbf{9.49} \\
    \midrule
    \multicolumn{6}{c}{1 Counterfact Editing on GPT-J (6B)} \\
    \midrule
    On Value & 100.00 & 98.18 & 6.04 & 16.15 & 0.79 \\
    \rowcolor{ROW_COLOR}
    On Key & 100.00 & \textbf{98.44} & \textbf{28.89} & \textbf{54.77} & 0.83 \\
    \midrule
    \multicolumn{6}{c}{1 zsRE Editing on GPT2-xl} \\
    \midrule
    On Value & 99.89 & 57.86 & 24.28 & 43.81 & 1.27 \\
    \rowcolor{ROW_COLOR}
    On Key & 99.89 & \textbf{69.90} & \textbf{24.75} & \textbf{46.35} & \textbf{0.88} \\
    \midrule
    \multicolumn{6}{c}{10 zsRE Editing on GPT2-xl} \\
    \midrule
    On Value & 99.89 & 98.11 & 19.36 & 41.75 & 0.68 \\
    \rowcolor{ROW_COLOR}
    On Key & 99.89 & 96.94 & \textbf{27.80} & \textbf{53.29} & 0.71 \\
    \midrule
    \multicolumn{6}{c}{1 zsRE Editing on GPT-J (6B)} \\
    \midrule
    On Value & 99.11 & 56.32 & 21.81 & 40.71 & 83.12 \\
    \rowcolor{ROW_COLOR}
    On Key & 98.57 & \textbf{69.19} & \textbf{24.64} & \textbf{46.02} & \textbf{12.63} \\
    \bottomrule
    \end{tabular}
    }
    \vskip -0.1in
\end{table}

Table~\ref{tab:more_edit} reports the average metrics of 20877 runs on counteract editing and 19086 runs on zsRE editing.
Updating $K$ achieves robust and considerable performance gains on the \textbf{Score}, especially since we can observe a large gap on the \textbf{Specificity}.
Although updating $V$ sometimes consumes slightly less time, it becomes largely inefficient when updating takes longer.
The results demonstrate the superiority of updating $K$ rather than $V$ to intervene in pre-trained transformer models about how it processes and stores information.

\subsection{More experimental results about the LoRA Tuning}
\label{sec:more_lora}

Our research compares the performance of different LoRA target settings in the more user-friendly context of 4-bit quantization. 
This approach is significant as it addresses the practical concerns of model size and computational efficiency, which are crucial for real-world applications.

Our findings consistently show that modifying keys outperforms modifying values.
This consistent superiority is notable, especially considering that the models with modified keys have fewer trainable parameters than those with modifications in q and v projections. 
Despite the reduced parameter count, the key-modified models still achieve better results.

\begin{table}[htbp!]
\small
    \vspace{-1.0em}
    \setlength{\tabcolsep}{4pt}
    \renewcommand{\arraystretch}{1.0}
    \centering
    \caption{\footnotesize LoRA instruction and multi-task tuning results under a 4-bit quantization setting}
    \vskip -0.07in
    \label{tab:lora_performance_4bit}
    \resizebox{0.9\textwidth}{!}{
    \begin{tabular}{l|c|cccccccc}
    \toprule
    \rowcolor{COLOR_MEAN}
    \textbf{Lora Target} &\textbf{Trainable} & \textbf{Bool-q} & \textbf{CB} & \textbf{20-news} & \textbf{Race} & \textbf{COPA} & \textbf{QNLI} & \textbf{6-Avg} \\
    \midrule
    \multicolumn{9}{c}{LoRA rank=8} \\
    \midrule
    q v & 0.062\% & 67.80 & 42.86 & 53.90 & 52.80 & 68.00 & 48.20 & 55.59 \\
    $\text{Value}_{\text{down}}$ & 0.057\% & 55.30 & 50.00 & 65.40 & 52.80 & 34.00 & 48.60 & 51.02 \\
    $\text{Key}_{\text{gate}}$ & 0.057\% & 71.60 & 55.71 & 68.50 & 53.00 & 63.00 & 49.30 & 60.19 \\
    \rowcolor{ROW_COLOR}
    $\text{Key}_{\text{up}}$ & 0.057\% & 71.30 & 60.71 & 61.00 & 55.80 & 70.00 & 54.30 & \textbf{62.19} \\
    \midrule
    q v $\text{Value}_{\text{down}}$ & 0.119\% & 55.40 & 41.07 & 50.80 & 58.10 & 58.00 & 50.00 & 52.23 \\
    q v $\text{Key}_{\text{gate}}$ & 0.119\% & 59.30 & 50.00 & 55.40 & 62.60 & 66.00 & 52.70 & 57.67 \\
    \rowcolor{ROW_COLOR}
    q v $\text{Key}_{\text{up}}$ & 0.119\% & 69.20 & 48.21 & 59.30 & 61.80 & 56.00 & 54.90 & \textbf{58.24} \\
    \midrule
    \multicolumn{9}{c}{LoRA rank=16} \\
    \midrule
    q v & 0.124\% & 70.20 & 41.07 & 61.30 & 56.70 & 63.00 & 51.40 & 57.28 \\
    $\text{Value}_{\text{down}}$& 0.115\% & 62.60 & 53.57 & 60.10 & 53.40 & 66.00 & 51.20 & 57.81 \\
    $\text{Key}_{\text{gate}}$ & 0.115\% & 72.10 & 46.43 & 60.20 & 56.10 & 64.00 & 59.90 & 59.79 \\
    \rowcolor{ROW_COLOR}
    $\text{Key}_{\text{up}}$ & 0.115\% & 68.30 & 50.00 & 60.20 & 57.40 & 74.00 & 52.70 & \textbf{60.43} \\
    \bottomrule
    \end{tabular}
    }
    \vskip -0.1in
\end{table}



\end{document}